\pdfoutput=1
\documentclass[11pt]{article}

\usepackage[review]{ACL2023}

\usepackage{times}
\usepackage{latexsym}

\usepackage[T1]{fontenc}
\usepackage[utf8]{inputenc}

\usepackage{microtype}

\usepackage{inconsolata}
\usepackage{amsmath}
\usepackage{mathtools}
\usepackage{xcolor}
\usepackage{enumitem}
\usepackage{comment}
\usepackage{graphicx}
\usepackage{multirow}
\usepackage{array}
\usepackage{caption}
\usepackage{fontawesome5}
\usepackage{hyperref}
\usepackage{wrapfig}
\usepackage{graphicx}
\usepackage{caption}
\usepackage{booktabs}
\usepackage{adjustbox}
\usepackage{footnote}
\makesavenoteenv{table}
\title{Explainability in Neural Networks for Natural Language Processing Tasks}

\author{
Melkamu Mersha\textsuperscript{1}, Mingiziem Bitewa\textsuperscript{2}, Tsion Abay\textsuperscript{3}, Jugal Kalita\textsuperscript{1}\\
\textsuperscript{1} University of Colorado Colorado Springs, CO, USA\\
\textsuperscript{2}University of Gondar, Gondar, Ethiopia\\
\textsuperscript{3}SOS HGS, Bardar, Ethiopia
}

\begin{document}
\makeatletter\acl@finalcopytrue
\maketitle

\begin{abstract} 
Neural networks are widely regarded as black-box models, creating significant challenges in understanding their inner workings, especially in natural language processing (NLP) applications. To address this opacity, model explanation techniques like Local Interpretable Model-Agnostic Explanations (LIME) have emerged as essential tools for providing insights into the behavior of these complex systems. This study leverages LIME to interpret a multi-layer perceptron (MLP) neural network trained on a text classification task. By analyzing the contribution of individual features to model predictions, the LIME approach enhances interpretability and supports informed decision-making. Despite its effectiveness in offering localized explanations, LIME has limitations in capturing global patterns and feature interactions. This research highlights the strengths and shortcomings of LIME and proposes directions for future work to achieve more comprehensive interpretability in neural NLP models.
\end{abstract}


% \begin{keyword}
% XAI\sep explainable artificial intelligence\sep interpretable deep learning\sep explainable machine learning\sep evaluation framework\sep evaluation metrics\sep large language models\sep LLMs\sep interpretability\sep natural language processing\sep NLP\sep explainability techniques\sep black-box models.
% \end{keyword}

\section{Introduction}
In today’s technology-driven world, natural language processing (NLP) has become a cornerstone of many applications, enabling systems such as voice assistants, email filters, chatbots, and predictive text to facilitate seamless communication \cite{young2018recent}. These advances, powered by neural network models, have revolutionized how machines understand and process human language. However, despite their remarkable success, neural network models are often considered opaque black-box systems, leaving users and researchers with limited insights into how these models generate their predictions. This lack of transparency raises critical concerns, particularly in high-stakes domains such as healthcare, finance, and law, where accountability and trust are paramount.
The concept of interpretability—defined as a model's ability to provide human-understandable reasoning for its predictions \cite{lipton2018mythos}—has emerged as a vital area of research. Understanding the inner workings of neural network models not only fosters trust but also aids in diagnosing errors, identifying biases, and ensuring ethical AI deployment. However, achieving interpretability for neural network-based NLP models is especially challenging due to the complexity of their layered architectures and the intricacies of language itself \cite{alishahi2019analyzing}.
To address these challenges, this study employs Local Interpretable Model-Agnostic Explanations (LIME), a widely used technique for explaining black-box models. LIME generates localized explanations by approximating a model’s behavior for individual instances, focusing on the contributions of specific features to predictions. This approach is particularly valuable for text classification tasks, as it allows for a granular understanding of how words influence a model's output. While LIME provides significant insights at the instance level, it fails to address global interpretability and capture interactions between features.
This study explores the application of LIME to a multi-layer perceptron (MLP) neural network trained on text classification, highlighting the technique's strengths and limitations. By shedding light on the decision-making process of neural NLP models, this study contributes to the broader goal of making AI models more transparent, trustworthy, and accountable.

\section{Background}
\subsection{Why Interpretability Matters}
In machine learning, model interpretability is a cornerstone for understanding how systems operate, particularly in real-world applications that influence critical decisions \cite{mersha2024explainable}. For example, in fields such as healthcare, finance, and legal systems, the impact of model predictions can be profound. However, traditional evaluation metrics like classification accuracy often fail to provide meaningful insights into a model's behavior and decision-making process \cite{doshi2017towards}. Interpretability helps ensure that models align with human expectations and plays a crucial role in identifying biases, uncovering vulnerabilities, and debugging unexpected outcomes. By making models more transparent and understandable, interpretability fosters trust, enhances accountability, and enables more informed decision-making.

\subsection{Explanation Methods}
Various explanation techniques have been developed to address the interpretability challenges in machine learning models. 
\subsubsection{Model-agnostic methods }
Model-agnostic methods are designed to work with any machine learning model, regardless of its complexity or structure. Techniques such as Local Interpretable Model-Agnostic Explanations (LIME), SHapley Additive exPlanations (SHAP), and perturbation analysis provide flexibility and adaptability by generating explanations that are independent of the underlying model architecture. These methods are beneficial for comparing different models where model architecture may not be fully accessible.
\subsubsection{Model-specific methods}
Model-specific methods, on the other hand, are tailored to specific types of models or architectures. For instance, Integrated Gradients and Attention mechanisms leverage the internal structure of neural networks to provide deeper insights into their decision-making processes. These methods often offer more fine-grained explanations but are limited to the models they are designed for, reducing their general applicability.
\subsubsection{Example-based methods}
Example-based methods rely on specific instances to explain model behavior. Techniques such as adversarial examples, influence functions, and counterfactual explanations highlight how a model responds to particular inputs or scenarios. These methods complement feature attribution approaches by providing alternative perspectives on model predictions and their underlying logic

\subsection{Local Interpretable Model-agnostic Explanation}
Local Interpretable Model-Agnostic Explanations (LIME) is a widely recognized technique for explaining black-box models by approximating their behavior locally around individual predictions \cite{yigezu2024ethio}. Instead of explaining the entire model globally, LIME focuses on generating localized, interpretable representations that make sense to humans. Its functionality varies depending on the type of data it is applied to. For text data, LIME creates explanations by selectively removing or masking words and analyzing their impact on predictions. For tabular data, it generates perturbations by altering feature values, while for images, it uses superpixel segmentation to modify specific regions. By training local surrogate models to mimic the behavior of the original black-box model, LIME provides instance-specific insights into how individual features contribute to predictions. Despite its advantages in localized interpretability, LIME's limitations include its inability to address global model behavior and its dependency on the quality of surrogate models.

1)	Text Data: 
New texts are generated by randomly removing words from the original text. The dataset is represented with binary features; the feature is 0 if the word is removed and 1 if the word is not removed.

\includegraphics[scale=0.41]{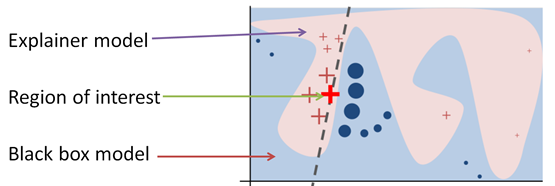}\\
\indent {\small Figure 1: LIME explanation logic \cite{ribeiro2016should}}\\

2)	Tabular Data: 
Tabular data is in table form data. The table columns represent the feature, and the rows represent the instances. \\
3)	Image Data:  
The functionality of LIME for images is different from the text and tabular data. First, LIME segments the image into “supperpixels” based on corresponding pixels with similar colors. Probabilistic prediction is made by turning superpixels on or off.

\section{Literature Review }
Neural networks with multiple layers can automatically learn complex patterns and representations from large datasets, with applications in computer vision, natural language processing (NLP), healthcare, and autonomous systems \cite{alshami2024smart}. Recent research has extensively focused on the interpretability and analysis of neural network models for natural language processing (NLP), addressing the critical challenge of understanding their complex decision-making processes. Interpretability has emerged as a cornerstone of explainable AI (XAI), particularly for NLP models deployed in high-stakes domains.

Lipton \cite{lipton2018mythos} conducted a seminal study that clarified the concept of interpretability in machine learning and underscored its significance for neural NLP models. The study established a foundational framework for defining interpretability, including taxonomies and evaluation criteria, highlighting its pivotal role in fostering trust and accountability in AI systems.

Ribeiro et al. \cite{ribeiro2016should} and Montavon et al. \cite{montavon2018methods} explored a range of neural network interpretation techniques, including Local Interpretable Model-Agnostic Explanations (LIME), Gradient-Based Sensitivity Analysis, and SHapley Additive exPlanations (SHAP). Ribeiro et al. emphasized LIME’s versatility across data modalities, while Montavon et al. advanced the understanding of gradient-based methods by leveraging their compatibility with deep neural network architectures.

Belinkov et al. \cite{belinkov2020interpretability} provided a comprehensive discussion on neural network interpretability, covering structural analyses through probing classifiers, behavioral evaluations via test suites, and interactive visualizations for debugging model behavior. Their work critically analyzed the strengths and weaknesses of various interpretability methods, offering insights into their applicability and limitations, such as the inability to capture contextual dependencies in language tasks fully.

Fan et al. presented a systematic review of interpretability techniques for neural networks, focusing on their applications in high-stakes domains like healthcare, finance, and legal systems. Their study also identified emerging trends in interpretability research, such as integrating XAI into decision-making pipelines and addressing biases in neural models. This review provided a roadmap for the practical implementation of interpretability frameworks, emphasizing the need for domain-specific adaptations.

Despite significant progress, existing studies highlight several challenges, such as balancing local and global interpretability, addressing model-specific nuances, and evaluating explanations against human reasoning benchmarks. These works collectively set the stage for advancing the field of XAI by integrating novel interpretability methods tailored for NLP tasks, ensuring models are not only accurate but also explainable and trustworthy.

% ///////////////////
% Much recent research has focused on interpreting and analyzing neural network models for natural language processing task. \cite{lipton2018mythos} presented a study on what is interpretability and why is it important to neural NLP models. \cite{ribeiro2016should} and \cite{montavon2018methods} explored various types of neural network interpretation techniques such as Local Interpretable Model, Gradient-Based Sensitivity Analysis, and SHAP. \cite{belinkov2020interpretability} discussed various aspects of neural network interpretation, such as structural analyses using probing classifiers, behavioral studies, test suites, and interactive visualizations, and highlighted the most applied analysis methods with their specific limitations and shortcomings of current approaches. \cite{fan2021interpretability} presented a systematic review of the interpretability of neural networks and described applications of interpretability in high stakeholders.

\section{Methodology}
The agnostic model explanation method is the most common explanation technique that can be applied to different types of ML models \cite{gholizadeh2021model}. Local Interpretable Model-agnostic Explanations (LIME) is an agnostic method \cite{ribeiro2016should}. LIME focuses on training local surrogate models to explain distinct predictions rather than training a global surrogate model.\\
LIME works on text, tabular data, and images. The LIME explainer turns on or off a single word or pixel in text and image data. In tabular data, the LIME explainer selects an instance being interpreted from the given text classification dataset and generates random perturbations around it to develop a new dataset. Then, it uses a neural network classifier to predict the class of the newly generated dataset. By computing the distance between the instance being interpreted and each perturbation, it finds the weight of the newly generated dataset. From this newly generated dataset, it uses its class prediction and weight to fit a linear model.  

\includegraphics[scale=0.55]{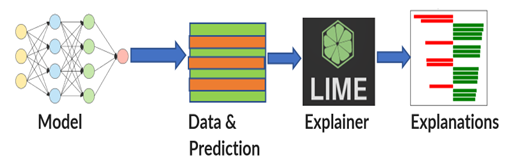}
        Figure. 2: System architecture 

\section{Implementation}
This project comprises two main implementation components: developing a text classification model and generating explainable insights using Local Interpretable Model-Agnostic Explanations (LIME). The model explanation is the primary focus, with the text classification component serving as the foundation for the interpretability analysis.
\subsection{ Text Classification}
Text classification, also known as text categorization, involves organizing textual data into predefined categories \cite{shah2019predictive}. This task is foundational in natural language processing (NLP) and finds applications in spam detection, sentiment analysis, topic labeling, and more.
For this study, a Multilayer Perceptron (MLP) was employed as the classifier due to its ability to learn complex patterns in textual data through non-linear transformations. The MLP architecture is particularly suited for classification tasks, offering flexibility and robustness when combined with effective preprocessing and feature engineering techniques.

% Text classification or text categorization is the process of categorizing text into organized groups \cite{shah2019predictive}.
\subsubsection{Dataset}
The 20 Newsgroups dataset, available at the UCI Machine Learning Repository (https://archive.ics.uci.edu/ml /machine-learning-databases/20newsgroups-mld/), is a benchmark dataset widely used in text categorization tasks. It comprises 20 categories such as sci.space, rec.sport.hockey, and talk.politics.mideast, with a total of 20,000 documents (1,000 per category). The dataset contains unstructured text with varying lengths and includes noise, such as headers, signatures, and quoted text.

For this study, 80\% of the data (16,000 documents) is used for training, and 20\% (4,000 documents) for testing.
Challenges: Overlapping vocabularies between categories and noisy metadata make classification challenging but realistic.
This dataset’s diversity and complexity make it ideal for evaluating the classification model and testing LIME-based explainability \cite{tonja2023first}.

% The 20newsgroups dataset is our input data for this classifier model (https://archive.ics.uci.edu/ml /machine-learning-databases/20newsgroups-mld/). It has twenty different categories and 20,000 instances.
\subsubsection{Prepossessing}
Preprocessing is a critical step in preparing textual data for analysis, ensuring consistency and reducing noise to improve model performance. The process began with case normalization, converting all text to lowercase to eliminate case sensitivity and reduce feature dimensionality. Next, excess whitespace between words, along with punctuation marks and special characters, was removed to focus on meaningful content. Stopword removal was applied to filter out frequently used but semantically insignificant words, such as "the," "is," and "and," using a predefined stopword list. The text was then tokenized into word-level tokens to prepare for feature extraction. To further enhance consistency, stemming and lemmatization were applied to reduce words to their root forms, with lemmatization prioritized for its contextual accuracy. Additionally, irrelevant metadata, such as email headers, signatures, and quoted text, was stripped from the documents to eliminate noise. Finally, text length filtering was performed to remove extremely short or excessively long documents, ensuring uniformity across the dataset. These preprocessing techniques collectively transformed the raw text into a structured format suitable for feature extraction and classification, optimizing the dataset for downstream tasks.

% \subsubsection{Tokenization}
% Tokenization is splitting paragraphs, sentences, or phrases into the smallest units \cite{huang2007rethinking}. All words in a document are split as single words. Word tokens are preferred for this text classification model development.  Word tokens give better results than character tokens \cite{choi2016text}. 

\subsubsection{Developing a Dictionary}
A dictionary was developed to serve as a repository of unique features extracted from the dataset, ensuring that only meaningful words contribute to the classification task. The process involved several key steps, as discussed in an earlier section. This dictionary contains a comprehensive and optimized set of unique features, enabling efficient text classification and enhancing the interpretability of LIME explanations by focusing on the most relevant tokens.

\subsubsection{Feature Engineering}
Feature engineering transforms the cleaned text data into a structured format suitable for modeling, where each unique word in the dictionary serves as a feature. The dictionary contains uniquely identified words and their frequency of occurrence in each document, forming the basis for feature extraction. Initially, the Bag-of-Words (BoW) approach was used to represent documents as sparse vectors, with each entry corresponding to the count of a specific word. Word2Vec embeddings were employed as an advanced feature representation method to capture semantic relationships and contextual information. Unlike BoW, Word2Vec generates dense vector representations of words, where words with similar meanings or contexts have closer embeddings in the vector space.

To further optimize the feature set, a document's average of word embeddings was calculated to create a single, fixed-length feature vector per document, enabling compatibility with the model input. Low-frequency words were excluded to reduce noise, and embeddings were normalized to ensure consistency across varying document lengths. This approach allows the model to leverage the semantic and contextual properties of words, enhancing its ability to generalize and improve predictions' interpretability through LIME explanations.

\subsubsection{Model Building}
To build an effective classifier, a Multilayer Perceptron (MLP) neural network was chosen for its ability to capture non-linear relationships in the data. The dataset, comprising features derived from Word2Vec embeddings and their corresponding category labels, was split into 80\% training and 20\% testing subsets to ensure a balance between model learning and performance evaluation. 
 Hyperparameters, including learning rate, batch size, and the number of epochs, were optimized through grid search to achieve the best performance. After training, the model's generalization ability was validated on the test dataset using accuracy, precision, recall, and F1 score metrics. This well-structured approach ensured the MLP classifier could effectively distinguish between categories while providing a solid foundation for generating explainability insights using LIME.

\subsection{ LIME Model Explanation}
The LIME model explanations use an interpretable data representation, not the actual features used by the classifier model, because the interpretable data representation is human-understandable. For example, a binary vector (bag of words) is an interpretable representation of the presence or absence of a word \cite{ribeiro2016should}.\\
The LIME output is a list explanation that tells the effect of each feature on the prediction of a given instance. This is the notion of local interpretability. Based on this local interpretability, we can also determine which feature has a high contribution to changing the prediction output. The LIME explanation is discussed as follows. 
We select an instance (document) by its index and feed it to the classifier model.
The neural networks classifier predicts the probability of a selected instance (document index 20) class being atheism is 0.42, and class Christian is 0.58 as shown from experiment 1. Now the question comes to a neural network,\textbf{ “Why should I trust you?"}.\\
If the predicted output of a classifier model does not fit what we expected from that selected instance, the classifier’s prediction cannot be trusted.
The LIME model generates the following output for the selected instances to answer this black box model question.

% \subsubsection{Train the LIME Model}
The dataset is split into 80\% as taring and 20\% as validation. Three significant parameters are required to train the model:  dataset with the label, number of features, and training samples.  
\includegraphics[scale=0.47]{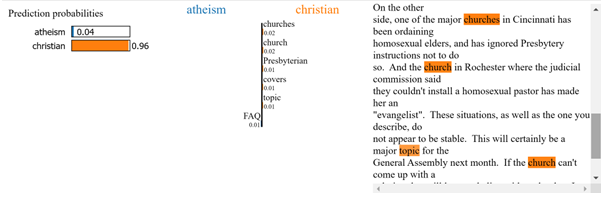}\\
\indent {\small Figure 3: LIME explanation for the input instance}\\

\includegraphics[scale=0.6]{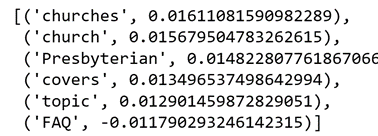}\\
\indent {\small Figure 3:the weight of features}\\
% \subsubsection{Validate the LIME Model}
The validation dataset is preprocessed with special attention because the 20newsgroup dataset is problematic in validating the model \cite{ribeiro2016should}. This preprocessing implementation is one of the contributions that enabled us to use the same dataset for training and validation.

\includegraphics[scale=0.44]{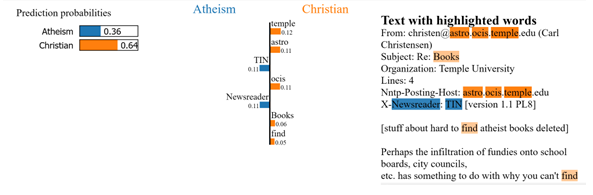}
{\small Figure 3: LIME explanation before preprocessing}

\includegraphics[scale=0.35]{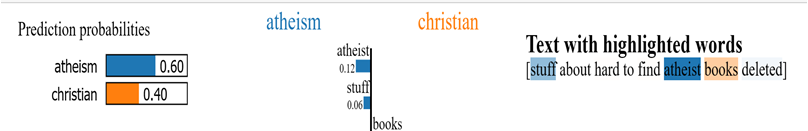}
{\small Figure 3: LIME explanation after preprocessing}
\\
\textbf{Experiment 1}\\
\\
\includegraphics[scale=0.65]{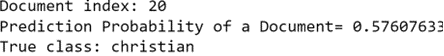}\\
{\small Figure 3: LIME explanation for the input instance}\\
\\
The LIME's explanation as a list of features (unique words) and prediction probability:
\includegraphics[scale=1]{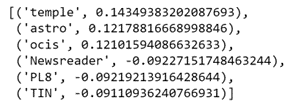}\\
{\small Figure 4: LIME list of explanation}\\
\\
\includegraphics[scale=0.7]{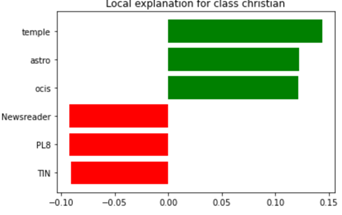}\\
{\small Figure 5: LIME explanation with plot visualization}\\
\\
\includegraphics[scale=0.7]{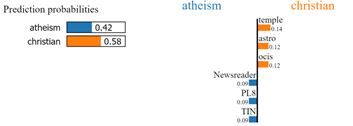}\\
{\small Figure 6: Prediction probability and local explanation for class christian and atheism}\\

\includegraphics[scale=0.9]{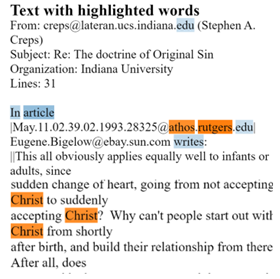}\\
{\small Figure 7: LIME explanation with highlighted output words}\\
\\
\textbf{Experiment 2}\\
\\
\includegraphics[scale=1]{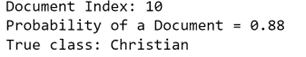}\\
{\small Figure 8: LIME explanation for an input instance}\\
\\
The LIME's explanation as a list of features (unique words) and prediction probability:
\includegraphics[scale=1]{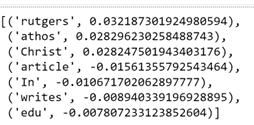}\\
{\small Figure 9: LIME list of explanation}\\
\\
\includegraphics[scale=0.7]{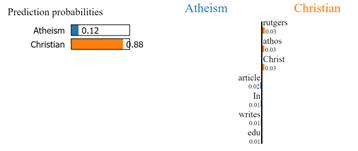}\\
{\small Figure 10: Prediction probability and local explanation for class christian and atheism}\\
According to experiment 1 and 2 outputs, we can find a basic understanding of how the neural networks model work. However, how is each selected feature has a high weight, and why does each word belong to that class? Unfortunately, LIME does not answer these questions.
\subsection{Analysis}
LIME has different parameters. The number of features is an essential parameter for this text model explanation task. Other parameters, such as dist\_fun and kernel\_width used for tabular data. Experiment 1 uses six features, and experiment 2 uses seven features. The number of features is determined depending on the input text size. Experiments 1 and 2 show that each feature(word) has a positive or negative value, which indicates a contribution of each word to the prediction process. For example, in experiment 2, the word 'rutgers' is the university's name. In actual human reasoning,  the word 'rutgers' does not make sense to the Christian class. However, it contributes high weight to the Christian class prediction; because the frequency of the word 'rutgers' in the Christian class determines this situation since a dictionary contains an unordered collection of words/features and that does not count a word's structure, meaning, and positioning in the document. \\
(NOTE: This analysis is not a final stage. There will be some future discussion and additional ideas)
\section{Future Work}
In this project, we utilized the Bag-of-Words (BoW) representation to explain the MLP classifier model. However, future work will focus on leveraging more sophisticated models to enhance explainability and performance. Specifically, adopting transformer-based models like BERT (Bidirectional Encoder Representations from Transformers) or Sentence Transformers can provide deeper contextual understanding and more accurate feature representations \cite{mersha2024semantic}.

Explaining predictions from transformer models presents new opportunities for interpretability research. Future efforts will explore techniques like Integrated Gradients, Layer-wise Relevance Propagation (LRP), and Attention-based Explanations to analyze the decision-making process of these complex architectures. Combining these methods with LIME could offer a more comprehensive view of both local and global interpretability.

Furthermore, we aim to integrate these explainability techniques into end-to-end NLP pipelines, focusing on applications where transparency is critical, such as healthcare, finance, and legal systems. Evaluating the scalability of these methods across diverse datasets and multilingual contexts will also be a key area of exploration. By transitioning to transformer-based models and state-of-the-art interpretability methods, future research can significantly advance our understanding of neural networks' behavior and improve their trustworthiness in real-world applications.

\section{Conclusion}
Explaining the decision-making processes of neural network models remains challenging, especially when compared to the straightforward nature of explaining linear models. This study demonstrated the application of Local Interpretable Model-Agnostic Explanations (LIME) as a powerful and versatile tool for interpreting neural network models in natural language processing (NLP). Using a Multilayer Perceptron (MLP) for text classification, LIME provided instance-specific insights by identifying and quantifying the contribution of individual features to the model's predictions.  The effectiveness of LIME in generating interpretable explanations hinges on the selection of features for each instance, which is influenced by the size and complexity of the input text. LIME produces a list of features with weighted values, clearly indicating how each feature impacts the prediction. This localized approach to interpretability supports model debugging, bias detection, and trust-building in neural network models, particularly in applications requiring accountability and transparency. However, the study also highlighted LIME's limitations, such as its reliance on feature selection strategies and inability to capture global model behavior or complex feature interactions. 
In conclusion, LIME provides a strong foundation for interpreting neural network models, but future work should focus on leveraging advanced architectures and explanation methods to address its limitations. By doing so, we can achieve more comprehensive and meaningful explanations, paving the way for more transparent and trustworthy AI systems in NLP and beyond.

% \bibliographystyle{apalike}
% \bibliography{references}

% \end{document}

\bibliography{anthology,custom}
\bibliographystyle{acl_natbib}

\end{document}